\documentclass[twoside,11pt]{article}

\usepackage[letterpaper,top=1in,bottom=1in,left=1.1in,right=1.1in]{geometry}
\usepackage{times}                 
\usepackage[T1]{fontenc}
\usepackage[utf8]{inputenc}

\usepackage{amsmath,amssymb,amsthm}
\usepackage{mathtools}
\usepackage[ruled,vlined]{algorithm2e}

\usepackage{graphicx}
\usepackage{booktabs}
\usepackage{multirow}
\usepackage{array}
\usepackage{tabularx}
\usepackage{xcolor}
\usepackage{subcaption}

\usepackage{tikz}
\usetikzlibrary{arrows.meta,positioning,fit,backgrounds,calc}

\usepackage[numbers,sort&compress]{natbib}

\usepackage[colorlinks=true,linkcolor=blue!55!black,citecolor=blue!55!black,%
            urlcolor=blue!55!black,breaklinks=true]{hyperref}
\usepackage{url}

\theoremstyle{plain}

\theoremstyle{definition}
\newtheorem{definition}{Definition}

\theoremstyle{remark}

\usepackage{xspace}
\newcommand{\method}{\textsc{HarnessEvo}\xspace}
\newcommand{\LOI}{\textsc{loi}\xspace}     
\newcommand{\LOO}{\textsc{loo}\xspace}     
\newcommand{\croleone}{\ensuremath{c_1}\xspace}
\newcommand{\cstrat}{\ensuremath{c_2}\xspace}
\newcommand{\cfmt}{\ensuremath{c_3}\xspace}
\newcommand{\cctrl}{\ensuremath{c_4}\xspace}

\begin{document}

\title{\bf Where Does Harness-Optimization Value Live? \\
       Localized Gains and the Budget-Splitting Trap in \\
       Self-Evolving LLM Agents}

\author{
  Michael Nguyen$^{4}$ \and Wei Chen Tan$^{1}$ \and Nurul Aisyah Hassan$^{2}$ \and Arvind Raman$^{3}$ \and Li Hua Lim$^{1}$ \and Ahmad Faiz Razak$^{2}$ \and Jia Hui Wong$^{3}$ \\[3pt]
  \normalsize $^{1}$Universiti Malaya \quad $^{2}$Universiti Sains Malaysia \quad $^{3}$Universiti Putra Malaysia \\
  \normalsize $^{4}$Monash University Malaysia \\
  \normalsize Kuala Lumpur / Penang / Serdang, Malaysia
}
\date{}
\maketitle

%

\begin{abstract}
A growing body of work makes a frozen large language model (LLM) into a better
agent by \emph{evolving its harness}---the textual scaffolding (persona, strategy,
format rules, control heuristics) wrapped around the model. The dominant recipe,
exemplified by reflective prompt evolution, treats the harness as a single
flat string and optimizes it as a whole. We ask a sharper question: \emph{where,
inside the harness, does the optimization value actually live?} We introduce
\method, which decomposes the harness into four named, separately-evolvable slots
(role, task-strategy, tool/format-rules, reflection/control) and tunes each one
in turn with the same reflective optimizer under an iso-budget, paired with a
leave-one-in / leave-one-out (\LOI/\LOO) protocol that attributes the held-out
gain to individual slots. We use this structured decomposition not as a
leaderboard method but as a \emph{microscope}. On ALFWorld with a frozen 7B
backbone, the full method \emph{ties} both baselines on the binary success
metric: \method reaches $0.657$ versus $0.642$ for stock and $0.642$ for
flat-string evolution (McNemar $p=0.617$ and $p=0.480$, both ns). The interesting
result is what the microscope reveals. (i) \emph{Localization:} the entire gain
is carried by one slot---reflection/control, whose leave-one-in gain is $+0.119$
($p=0.0046$, $22$ vs $6$ discordant tasks), while role, strategy and format are
individually null. (ii) \emph{A budget-splitting trap:} the structured method is
strongly sub-additive (full gain $+0.015 \ll \sum_{\text{slots}}$ leave-one-in
$+0.164$) because the reflective optimizer has an accept-and-rescore \emph{floor};
splitting an iso-budget of $64$ rollouts across four slots leaves $16$ per slot,
below that floor, so every slot froze at its empty seed and the load-bearing slot
was never captured. (iii) \emph{A budget-allocation prescription:} concentrating
the same budget on the high-credit slot recovers the discarded gain---a targeted
single-slot run reaches $0.761$ ($+0.119$, $p=0.0046$) at \emph{half} the split's
budget, and an all-to-control run at the full budget reaches $0.724$ ($+0.082$,
$p=0.0725$, borderline; we report both honestly and do not overclaim at the
larger budget). (iv) \emph{Task-contingency:} the phenomenon is benchmark-specific;
on WebShop every slot froze empty and all methods tie ($0.518$/$0.539$/$0.545$,
ns), and a targeted control run there stays null---confirming the WebShop result
is a genuine absence of recurrent, verbalizable control failures, not budget
starvation. The harness-optimization value that \emph{does} survive is
environment-grounded self-correction (``skip a repeated action that returned
\emph{Nothing happens}; open closed receptacles before taking; confirm the object
is held before placing; stop when the goal is satisfied''), not reward-exploiting
strategy. We conclude that harness value is localized, that uniform budget-splitting
is actively harmful, and that credit assignment plus budget concentration should
precede any structured agent-evolution scheme.
\end{abstract}

\vspace{0.5em}
\vspace{1em}

\begin{figure}[t]
  \centering
  \IfFileExists{figures/fig_teaser.pdf}{%
    \includegraphics[width=0.92\linewidth]{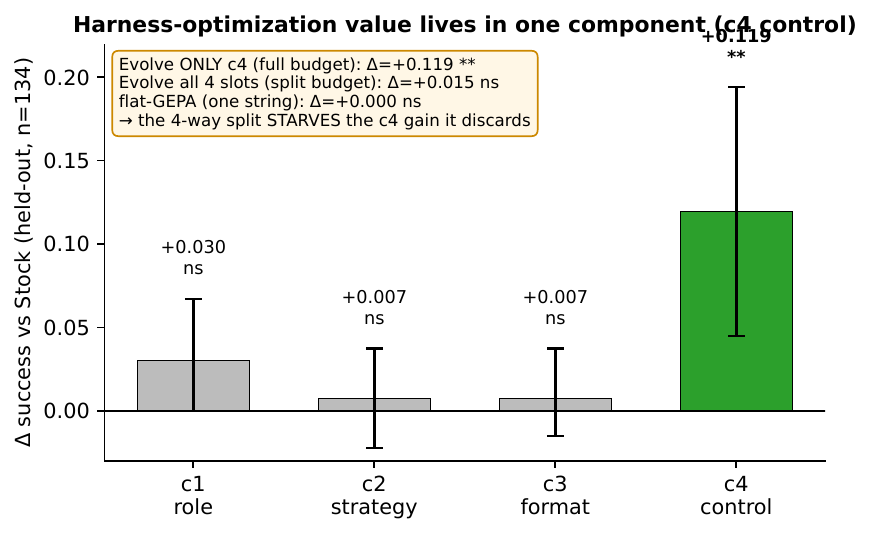}%
  }{%
    \fbox{\parbox[c][3.5cm][c]{0.92\linewidth}{\centering\itshape
      figures/fig\_teaser.pdf not found --- placeholder.}}%
  }
  \caption{\textbf{Harness-optimization value is concentrated in one component.}
    Per-slot leave-one-in (\LOI) gain on held-out ALFWorld ($n{=}134$, frozen 7B
    backbone): evolving \emph{only} the reflection/control slot (\cctrl) yields
    $+0.119$ ($p{=}0.0046$), while role (\croleone), strategy (\cstrat) and
    format (\cfmt) are individually indistinguishable from the stock harness.
    Evolving all four slots under the \emph{same} budget split four ways
    ($\Delta{=}+0.015$, ns) and flat single-string evolution ($\Delta{=}+0.000$,
    ns) both tie the baselines: the uniform split \emph{starves} the one slot
    that carries the gain. Error bars are $95\%$ bootstrap intervals on the
    paired held-out difference.}
  \label{fig:teaser}
\end{figure}

\section{Introduction}
\label{sec:intro}

A frozen large language model can be turned into a markedly stronger agent without
touching a single weight, simply by improving the text that surrounds it. The
agent's \emph{harness}---the system preamble, persona description, task strategy,
action templates and loop-control instructions injected around the model at every
step---determines, to first order, how an off-the-shelf LLM behaves in an
interactive environment. A rapidly growing literature exploits this fact: rather
than fine-tune, it \emph{optimizes the harness}, treating prompt text as a
parameter and searching over it with reflective feedback, evolutionary mutation,
or memory accumulation \citep{zhou2023ape,yang2024opro,fernando2023promptbreeder,
khattab2024dspy,agrawal2025gepa,wang2024awm,ace2025}. The flagship of this family,
reflective prompt evolution \citep{agrawal2025gepa}, reports that evolving a
single instruction string can match or beat reinforcement learning at a fraction
of the rollout cost. These methods are attractive precisely because they are
cheap, weight-free, and require only black-box access to the model.

Almost all of this work, however, optimizes the harness as an \emph{undifferentiated
whole}: a single flat string that is mutated, scored, and selected in one piece.
This conflates components that play very different roles. A persona line (``you
are a careful household assistant''), a task-decomposition strategy (``first
locate the object in the receptacles where it usually lives''), an action-format
rule (``emit one action per line using the template \texttt{take <obj> from
<recep>}''), and a control heuristic (``never repeat an action that returned
\emph{Nothing happens}; stop once the goal is satisfied'') are all swept into the
same string and optimized jointly. When such a method works, we learn \emph{that}
the harness can be improved; we do not learn \emph{which part of it} mattered, and
we therefore cannot say whether the next harness-optimization effort should spend
its budget on the persona, the strategy, the format, or the control logic. The
field has a leaderboard but no map.

This paper asks the map-making question directly: \textbf{where, inside the
harness, does the optimization value live?} We answer it with a deliberately
structured instrument. We decompose the harness into four named, mutually
exclusive, separately-evolvable slots---\croleone\ role/persona, \cstrat\
task-strategy, \cfmt\ tool/format-rules, \cctrl\ reflection/control-heuristics---and
evolve them one at a time by coordinate ascent, reusing an existing reflective
prompt optimizer slot-by-slot under a fixed total rollout budget that exactly
matches the budget given to flat single-string evolution. We pair this with a
leave-one-in / leave-one-out (\LOI/\LOO) credit-assignment protocol: \LOI\ evolves
a single slot alone and measures its marginal value over the stock harness, while
\LOO\ evolves everything except one slot and measures how much the full harness
loses without it. \emph{We stress at the outset that the structured decomposition
is the microscope, not the product.} We do not claim that decomposing the harness
yields a better agent; on our binary metric it does not. We claim that it lets us
see something the flat methods cannot.

What it sees is a striking, and in our view consequential, asymmetry. On ALFWorld
\citep{shridhar2021alfworld} with a frozen 7B backbone, the full structured method
ties both baselines---it reaches $0.657$ held-out success against $0.642$ for the
stock harness and $0.642$ for flat-string reflective evolution, and neither
difference is significant (McNemar $p=0.617$ versus stock, $p=0.480$ versus
flat). A reader who saw only this number would conclude, reasonably, that
structured harness evolution does nothing. But the \LOI\ map shows that the gain
is not absent---it is \emph{localized}. Evolving only the reflection/control slot
\cctrl\ lifts held-out success by $+0.119$ ($p=0.0046$; $22$ of the $28$ tasks on
which the two harnesses disagree flip in favour of the evolved control), whereas
evolving only the role, strategy, or format slot is individually null
($+0.030$, $+0.007$, $+0.007$; all ns). The value of harness optimization, at
least on this benchmark, lives almost entirely in one of the four components.

Why, then, does the full method tie? Because the structured method is strongly
\emph{sub-additive}, and we identify the mechanism. The full harness's gain over
stock is only $+0.015$, while the sum of the four \LOI\ gains is $+0.164$---an
order of magnitude larger. The cause is an interaction between coordinate-ascent
budget-splitting and a structural property of the reflective optimizer: it has an
accept-and-rescore \emph{floor}. Each optimizer iteration spends rollouts scoring
a parent and a mutated child on a minibatch, and only when a child is accepted
does it spend a further block of rollouts re-scoring the accepted child on the
full validation set. Splitting an iso-budget of $64$ rollouts evenly across four
slots leaves $16$ per slot---below the floor needed to accept and freeze even one
mutation---so the structured method froze \emph{every} slot at its empty seed,
producing a harness that is byte-identical to stock and never captured \cctrl's
gain at all. Naive structured evolution is thus \emph{worse than useless} here: it
is worse than a targeted single-slot run, because decomposition dilutes the budget
below the optimizer's working point. We call this the \emph{budget-splitting
trap}.

The trap immediately implies its own cure: under a fixed budget, one should not
split it uniformly but instead \emph{concentrate} it on the high-credit slot. We
verify this in two
ways and report both honestly. A leave-one-in run on the control slot at \emph{half}
the split's budget ($32$ rollouts, all on \cctrl) reaches $0.761$, a $+0.119$ gain
over stock that is significant ($p=0.0046$). An all-to-control run at the full
budget ($64$ rollouts) reaches $0.724$, a $+0.082$ gain that is positive and well
above the four-way split's $0.657$ but borderline by McNemar ($21$ vs $10$
discordant, $p=0.0725$, ns). The \emph{direction} is robust---both targeted runs
beat the split, and both clear it comfortably---but the exact magnitude is noisy
at $n=134$ given evolution and decoding stochasticity, so we make the deliberately
conservative claim that targeted single-slot evolution recovers a $+0.08$ to
$+0.12$ gain that uniform budget-splitting discards, and we do \emph{not} assert
significance at the larger budget.

Finally, the whole phenomenon is \emph{task-contingent}. On WebShop
\citep{yao2022webshop} with the same backbone and protocol, every slot froze at
its empty seed, the full method and both baselines tie ($0.518$ / $0.539$ /
$0.545$ dense score, all ns), and---crucially---a targeted all-to-control run there
\emph{also} stays null, with the control slot again freezing empty. This rules out
budget starvation as the explanation on WebShop: the optimizer had the
concentrated budget and still found nothing to accept. The difference between the
benchmarks is legible in the audited slot text. The control rules that help on
ALFWorld encode recurrent, verbalizable control failures---repeated dead-end
actions, closed receptacles, premature stopping---that recur across household
tasks; WebShop's purchasing task, with its dense attribute-match reward, simply
does not exhibit a control failure mode that a general heuristic can fix. The
value that harness optimization captures, where it captures anything, is
environment-grounded \emph{self-correction}, not reward-exploiting strategy---a
point we develop as a constructive counterpoint to concerns about optimizing
agents against gameable environment rewards.

\paragraph{Contributions.} We make four contributions, each stated as an honest
characterization rather than a leaderboard claim.

\begin{enumerate}
\item \textbf{A component-decomposed harness and a leave-one-in credit-assignment
protocol} that localize \emph{where} harness-optimization value lives. On ALFWorld
the value is overwhelmingly in the reflection/control slot (\LOI\ $+0.119$,
McNemar $p=0.0046$, $22$ vs $6$); role, strategy and format are individually null.
We do not claim a binary win---the full harness ties both stock and flat-string
evolution ($0.657$ / $0.642$ / $0.642$, ns).

\item \textbf{A sub-additivity result with an identified mechanism.} The full
gain ($+0.015$) is far below the sum of the four single-slot gains ($+0.164$). The
cause is an accept-and-rescore floor in coordinate-ascent reflective evolution:
an iso-budget of $64$ rollouts split four ways gives $16$ per slot, below the
floor, so the structured method froze every slot empty and never captured the
load-bearing gain. Uniform structured evolution is worse than a targeted run
because decomposition dilutes the budget below the optimizer's working point.

\item \textbf{A budget-allocation prescription.} Under a fixed budget,
concentrating it on the high-credit slot recovers the gain that uniform splitting
discards: a control-only run wins by $+0.119$ ($p=0.0046$) at \emph{half} the
split's budget, and an all-to-control run at the full budget confirms the
direction ($0.724$, $+0.082$, borderline). We report the significant ($@32$) and
borderline ($@64$) replications side by side and do not overclaim the larger
budget.

\item \textbf{A task-contingency boundary.} Localization is benchmark-specific.
On WebShop the method yields a full null---every slot froze empty
($0.518$/$0.539$/$0.545$, ns)---and a targeted control run there stays null too,
showing the WebShop result is a genuine absence of recurrent verbalizable control
failures, not budget starvation. We characterize \emph{when} harness optimization
helps (tasks with recurrent, verbalizable control failures: dead-end actions,
ordering constraints, premature stopping) via the audited control-rule text.
\end{enumerate}

Figure~\ref{fig:teaser} previews the central finding. Section~\ref{sec:related}
situates the work; Section~\ref{sec:problem} formalizes the harness-optimization
problem and the component decomposition; Section~\ref{sec:method} presents \method,
its architecture, the coordinate-ascent algorithm, and the \LOI/\LOO\ protocol;
Section~\ref{sec:setup} gives the experimental setup; Section~\ref{sec:results}
reports the main results; Section~\ref{sec:analysis} analyzes credit assignment
and the budget-splitting trap; Sections~\ref{sec:discussion}
and~\ref{sec:conclusion} discuss implications and limitations.

\section{Related Work}
\label{sec:related}

\subsection{Prompt and agent self-evolution}
\label{sec:rel:evolution}

The premise that a frozen model can be improved by optimizing its surrounding text
underlies a large and fast-moving literature. Early automatic prompt search treats
the instruction as a discrete object to be proposed and selected: APE generates and
ranks instructions with the model itself \citep{zhou2023ape}, OPRO casts the LLM as
an optimizer that iteratively proposes higher-scoring prompts from a trajectory of
past attempts \citep{yang2024opro}, and Promptbreeder evolves both task prompts and
the mutation prompts that produce them \citep{fernando2023promptbreeder}. Program-of-prompts
frameworks compile multi-stage LLM pipelines and tune their prompts as parameters:
DSP and DSPy formalize declarative LLM programs and optimize their instructions and
demonstrations \citep{khattab2022dsp,khattab2024dspy}. A related, test-time strand
adapts the prompt or scratchpad on the fly---dynamic cheatsheets that accumulate
reusable hints across queries \citep{suzgun2025dc} and scaling test-time compute by
revising the model's own intermediate text \citep{snell2024testtime,muennighoff2025s1}.
In all of these the optimized object is the prompt text, but it is optimized as one
piece; none asks which span of that text carries the value.

The work most directly relevant to ours, and the one we treat as the principal
foil, is reflective prompt evolution---GEPA \citep{agrawal2025gepa}---which evolves
a prompt by reflecting on rollout traces, mutating a single string, gating
candidates on a minibatch, and maintaining a Pareto frontier over instances.
\citet{agrawal2025gepa} show this flat-string scheme can outperform reinforcement
learning at far lower rollout cost. Our method reuses exactly this optimizer, but
applies it slot-by-slot rather than to one undifferentiated string; our finding is
that the value such an optimizer extracts is concentrated in a single component,
and that the structured iso-budget variant ties and can underperform the flat one
\emph{unless} the budget is localized to the load-bearing slot. We thus inherit
GEPA's mechanism and ask a question it was not designed to answer: not how much the
harness can improve, but which part of it carries the improvement.

A complementary strand evolves \emph{context} rather than instructions. Agent
Workflow Memory induces reusable workflows from experience and injects them as
context \citep{wang2024awm}; Agentic Context Engineering grows and curates an
evolving context for self-improving models \citep{ace2025}; memory architectures
such as MemGPT and MemoryBank manage long-horizon agent state
\citep{packer2023memgpt,zhong2024memorybank}. Skill- and experience-induction
methods---Voyager's skill library \citep{wang2023voyager}, ExpeL's experiential
learning \citep{zhao2024expel}, and the broader self-improvement line in which a
model bootstraps from its own outputs \citep{huang2022selfimprove,zelikman2022star,
muennighoff2025s1}---accumulate reusable knowledge across episodes. A parallel line
closes the loop through self-generated preference signals: self-rewarding and
meta-rewarding language models judge and improve their own responses
\citep{yuan2024selfrewarding,wu2024metarewarding}, and generative agents simulate
believable behaviour from accumulated memory \citep{park2023generative}. Self-evolving \emph{tool-use} optimization,
which assigns blame to individual tool calls and mutates a tool-use policy
accordingly, is closely related in spirit to our credit-assignment goal but
operates at the granularity of tool invocations rather than harness components
\citep[\emph{cf.}][whose blame-aware mutation localizes value at the tool-call
level, where our \LOI\ map localizes it at the harness-component
level]{yang2026evotool}. Surveys of LLM-agent autonomy and self-improvement map
this space \citep{su2025autonomysurvey,chen2024rplasurvey}. Across all of these,
the unit of optimization is a monolith---a string, a context, a skill set, a
policy---and the question of \emph{which sub-part of the harness carries the value}
is, to our knowledge, not posed. Our contribution is to pose and answer it, and to
show that the answer changes how a fixed optimization budget should be spent.

\subsection{LLM-agent frameworks and interactive benchmarks}
\label{sec:rel:agents}

The harness we decompose is, concretely, the scaffolding of a reasoning-and-acting
agent. ReAct interleaves chain-of-thought reasoning with environment actions
\citep{yao2023react} and is the loop our agent runs; Reflexion adds verbal
self-reflection across attempts \citep{shinn2023reflexion}; Tree-of-Thoughts and
self-consistency structure or aggregate the reasoning itself
\citep{yao2023tot,wang2023selfconsistency}. Self-correction and self-refinement
methods let a model critique and revise its own outputs
\citep{madaan2023selfrefine,gou2023critic,huang2024selfcorrect}, though their
reliability without external feedback is contested
\citep{kamoi2024selfcorrection}; our control slot is, in effect, a frozen,
distilled, environment-grounded version of such self-correction, evolved offline
rather than invoked online. Tool-use agents extend this to external APIs
\citep{qin2023toolllm}, and structured tool-exploration schemes organize the
search over tool calls \citep{yang2026tooltree}; retrieval-augmented generation
similarly equips an LLM with external knowledge at inference
\citep{lewis2020rag,guu2020realm,asai2024selfrag}, a context-side complement to the
prompt-side scaffolding we study.

Interactive benchmarks ground these agents. ALFWorld aligns embodied household
tasks with text-based TextWorld counterparts, requiring multi-step manipulation
under partial observability \citep{shridhar2021alfworld}; WebShop poses
instruction-conditioned web shopping with a dense attribute-match reward
\citep{yao2022webshop}; WebArena and GAIA push toward realistic web and
general-assistant tasks \citep{zhou2024webarena,mialon2023gaia}; AgentBench
\citep{liu2024agentbench}, $\tau$-bench \citep{yao2024taubench}, and SWE-bench
\citep{jimenez2024swebench} broaden the evaluation surface. We use ALFWorld and
WebShop precisely because they differ in the property our analysis turns out to
hinge on: ALFWorld exhibits recurrent, verbalizable control failures (repeated
dead-end actions, closed receptacles, premature stopping) that a general heuristic
can fix, whereas WebShop's purchasing task does not, which is exactly why
harness optimization helps on the former and is null on the latter.

\subsection{Credit assignment, ablation, and component analysis}
\label{sec:rel:credit}

Our central instrument is credit assignment: attributing an aggregate gain to the
parts of a system. Ablation is the standard tool, but it is usually applied to
architectural modules or training-signal components rather than to the slots of an
agent's prompt harness. Analyses of individual-unit importance show that single
directions or units can carry disproportionate functional weight in a network
\citep{morcos2018singledirections}, a structural analogue of our finding that one
of four harness slots carries essentially all the value. Decomposition-and-recombination
strategies such as branch-solve-merge \citep{saha2024branchsolvemerge} and the
generate-then-rank pattern \citep{li2024fromgeneration} restructure LLM computation
into parts, but do not measure per-part marginal value under a shared budget. A
separate question is what value the frozen model already realizes \emph{internally}:
work framing in-context learning as implicit optimization
\citep{vonoswald2023icgd,akyurek2023icllinear} helps explain why slots that merely
restate capabilities the model already has (persona, format) add nothing in our
map, while a slot that supplies missing control structure does. Our
leave-one-in / leave-one-out protocol is, deliberately, the simplest possible
credit-assignment design that isolates a slot's marginal contribution (\LOI) and
its removal cost (\LOO) on held-out tasks with paired significance tests, and our
sub-additivity result---full gain far below the sum of single-slot gains---is a
quantitative statement about how badly per-part value composes when the optimizer's
budget is split.

\subsection{Reward over-optimization, Goodhart effects, and grounded value}
\label{sec:rel:goodhart}

Optimizing an agent against a fixed objective invites the failure mode in which the
proxy is gamed rather than the goal achieved. This is the classical Goodhart
phenomenon \citep{goodhart1984problems}, formalized for reward learning as reward
over-optimization and hacking: as optimization pressure increases, measured reward
rises while true performance eventually falls \citep{gao2023scaling,
skalse2022defining,pan2022effects,amodei2016concrete}. This concern has shaped
preference-based fine-tuning, from learning rewards for summarization
\citep{stiennon2020summarize} to direct preference optimization and its AI-feedback
variants \citep{rafailov2023dpo,lee2024rlaif}. The risk is acute when an
agent is tuned against an automatically computed environment reward, and broad
treatments of agent misalignment catalogue when such optimization induces
specification-gaming behaviour \citep[\emph{e.g.}][which surveys the taxonomy and
causes of optimizing agents against environment reward, and motivates why a
\emph{gameable} dense reward is exactly where a harness optimizer should decline to
overfit]{yang2026misalignment}. Two aspects of our results speak to this concern as
a constructive counterpoint. First, the harness value that survives on ALFWorld is
\emph{environment-grounded self-correction}---rules that avoid repeated dead-end
actions and verify preconditions---rather than a strategy that exploits the binary
success checker; it generalizes across held-out task types because it fixes a real,
recurrent control failure. Second, on WebShop, whose dense reward is the more
gameable of the two, the optimizer declines to fill any slot even when handed a
concentrated budget: it finds no general control rule worth accepting, and the
method returns a genuine null rather than an over-fit strategy. In both cases the
optimizer captures grounded correction where it exists and abstains where it does
not, which is the behaviour one would hope for from a harness optimizer used in the
presence of an imperfect reward. Related lines on constitutional and preference-based
alignment \citep{bai2022constitutional,ouyang2022instructgpt} and on LLM-as-judge
evaluation---both its use \citep{zheng2023judging,liu2023geval,kim2024prometheus}
and its known biases \citep{gu2024judgesurvey,panickssery2024selfpref}---address the
reward side of this problem, since a harness optimizer is only as trustworthy as
the signal it optimizes; our contribution is on the optimization side,
characterizing what a black-box harness optimizer does and does not extract from a
given signal. Because our ALFWorld reward is an exact environment verifier rather
than a learned judge, the value we localize is not an artifact of a gameable scorer.

\subsection{Relation to recursive self-evolution}
\label{sec:rel:recursive}

A distinct axis of agent self-improvement is \emph{recursion}: agents (or
multi-agent systems) that revise their own improvement procedure, not merely their
prompt or context \citep{yang2026recursivemas,zou2025latentmas}. Recursion is
orthogonal to the question we study. We hold the optimizer fixed and ask where, in
a single non-recursive pass, its value lands inside the harness; we do not iterate
the meta-procedure. We make this distinction in prose to position the work, but our
empirical claims concern only single-pass, coordinate-ascent harness optimization,
and our budget-splitting analysis would apply to any scheme---recursive or not---that
distributes a fixed rollout budget across named harness components before the
optimizer's accept-floor is met.

\section{Problem Formulation}
\label{sec:problem}

\paragraph{Frozen-model agent.} We study an interactive agent built from a
\emph{frozen} language model $M$ (no gradient updates) acting in an environment
$\mathcal{E}$ over discrete steps. At each step the agent observes a textual
state, emits an action, and receives a new observation; an episode terminates with
an environment-verified outcome. The model's behaviour is conditioned on a textual
\emph{harness} $H$ that is injected at a single, fixed point in the agent's prompt
construction---the same point a ReAct agent uses for its strategy preamble
\citep{yao2023react}. Crucially, $H$ enters only as text: it never alters the
rollout \emph{code path} (decoding temperature, environment, action parser,
verifier are all fixed). Two harnesses that render to the same string produce
byte-identical rollouts.

\begin{definition}[Harness-optimization problem]
\label{def:problem}
Let $\rho(H)$ denote the expected held-out success of the frozen-model agent when
injected with harness text $H$, estimated by rolling out a held-out task set under
a deterministic ($\text{temperature}{=}0$) policy. Given a rollout budget $B$
(the number of environment rollouts the optimizer may spend during search), the
harness-optimization problem is to find
$H^\star \in \arg\max_{H} \rho(H)$ using at most $B$ rollouts, with $H$ selected on
a held-out validation set and reported on a disjoint test set.
\end{definition}

\paragraph{The stock harness.} The neutral baseline is the \emph{stock} harness
$H_0$, which carries no task guidance and renders to the empty string. Injecting
$H_0$ is byte-identical to running the agent with no preamble at all. Every
optimization method must \emph{earn} the content of $H$ through search starting
from this empty seed; this makes flat single-string evolution and our structured
method fair to compare, since both begin from $\rho(H_0)$ and spend the same budget
$B$.

\paragraph{Component decomposition.} We decompose the harness into four named,
mutually exclusive slots in a fixed render order,
\begin{equation}
H = (\croleone, \cstrat, \cfmt, \cctrl), \qquad
\mathrm{render}(H) = \croleone \,\Vert\, \cstrat \,\Vert\, \cfmt \,\Vert\, \cctrl,
\label{eq:render}
\end{equation}
where $\Vert$ concatenates non-empty slots (each under a short neutral section
header) and drops empty ones, so that $\mathrm{render}(H_0)=\varepsilon$ (the empty
string). The four slots, with disjoint semantic scope, are:
\begin{itemize}
\item \croleone\ \textbf{role / persona}: who the agent is and the disposition it
adopts (careful, methodical, goal-driven). No action templates or stop rules.
\item \cstrat\ \textbf{task-strategy}: the high-level approach for decomposing and
solving a task (what to do first, how to search the environment). No persona,
syntax, or loop heuristics.
\item \cfmt\ \textbf{tool / format-rules}: the exact action templates and output
format the agent must emit, and any hard syntactic rules. Copy-usable; no strategy
or control logic.
\item \cctrl\ \textbf{reflection / control-heuristics}: loop discipline and
self-correction rules (never repeat a failed action, verify before finishing, when
to backtrack, when to stop), optionally with a small structured control-knob line.
No persona, strategy, or low-level syntax.
\end{itemize}

\paragraph{Credit-assignment objective.} Beyond finding $H^\star$, our goal is
to \emph{attribute} the gain $\rho(H^\star)-\rho(H_0)$ to the four slots.
For a slot $c_k$ we define two held-out quantities, computed by evolving the
relevant slot(s) and evaluating on the test set:
\begin{align}
\Delta^{\mathrm{LOI}}_k &= \rho\big(\mathrm{LOI}_k\big) - \rho(H_0),
  &\text{(marginal value of $c_k$ alone over stock)}\\
\Delta^{\mathrm{LOO}}_k &= \rho\big(\mathrm{LOO}_k\big) - \rho(H_{\mathrm{full}}),
  &\text{(loss when $c_k$ is removed from the full harness)}
\end{align}
where $\mathrm{LOI}_k$ keeps only slot $c_k$ evolved (others at the empty seed) and
$\mathrm{LOO}_k$ evolves all slots except $c_k$. A value that is \emph{localized}
to slot $c_k$ exhibits a large positive $\Delta^{\mathrm{LOI}}_k$ with the other
slots' \LOI\ gains near zero. \emph{Sub-additivity} is the statement
$\rho(H_{\mathrm{full}})-\rho(H_0) \;\ll\; \sum_k \Delta^{\mathrm{LOI}}_k$, i.e.\
the jointly optimized harness captures far less than the sum of the individually
optimized slots. Table~\ref{tab:notation} collects the notation.

\begin{table}[t]
\centering
\small
\caption{Notation.}
\label{tab:notation}
\begin{tabularx}{\linewidth}{@{}lX@{}}
\toprule
Symbol & Meaning \\
\midrule
$M$ & frozen language-model backbone (Qwen2.5-7B; no weight updates) \\
$\mathcal{E}$ & interactive environment (ALFWorld or WebShop) \\
$H$ & harness: text scaffolding injected at the single fixed injection point \\
$H_0$ & stock harness (all slots empty; renders to $\varepsilon$; byte-identical to no-preamble) \\
$\croleone,\cstrat,\cfmt,\cctrl$ & the four slots: role, task-strategy, tool/format-rules, reflection/control \\
$H_{\mathrm{full}}$ & harness with all four slots evolved by coordinate ascent \\
$\rho(H)$ & expected held-out test success (or dense score) of the agent with harness $H$ \\
$B$ & rollout budget the optimizer may spend during search (iso-budget across methods) \\
$\mathrm{LOI}_k$ & leave-one-in: only slot $c_k$ evolved, others at empty seed \\
$\mathrm{LOO}_k$ & leave-one-out: all slots except $c_k$ evolved \\
$\Delta^{\mathrm{LOI}}_k$, $\Delta^{\mathrm{LOO}}_k$ & held-out marginal value of $c_k$ over stock / loss on removing $c_k$ \\
\bottomrule
\end{tabularx}
\end{table}

\section{Method: \method}
\label{sec:method}

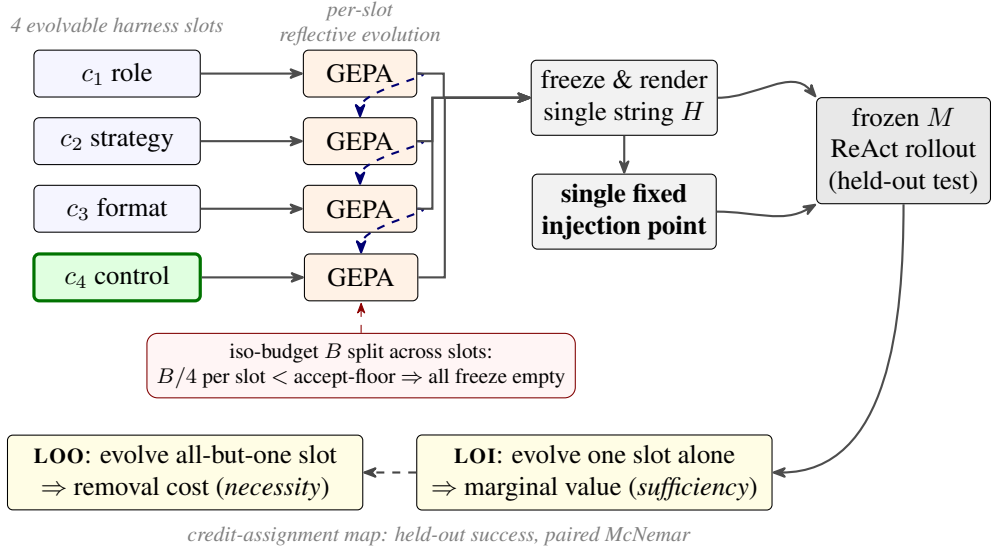
\begin{figure}[t]
\centering
\begin{tikzpicture}[
  font=\small,
  >={Stealth[round]},
  slot/.style={draw, rounded corners=2pt, minimum width=2.2cm, minimum height=0.62cm, align=center, fill=blue!4},
  ctrl/.style={slot, fill=green!12, draw=green!45!black, very thick},
  gepa/.style={draw, rounded corners=2pt, minimum width=1.5cm, minimum height=0.62cm, align=center, fill=orange!10},
  inj/.style={draw, rounded corners=2pt, minimum width=2.3cm, minimum height=0.85cm, align=center, fill=gray!10},
  cred/.style={draw, rounded corners=2pt, minimum width=4.7cm, minimum height=0.95cm, align=center, fill=yellow!12},
  lbl/.style={font=\scriptsize\itshape, gray},
  arr/.style={->, thick, gray!60!black},
]
\node[slot] (s1) {$c_1$ role};
\node[slot, below=0.26cm of s1] (s2) {$c_2$ strategy};
\node[slot, below=0.26cm of s2] (s3) {$c_3$ format};
\node[ctrl, below=0.26cm of s3] (s4) {$c_4$ control};
\node[lbl, above=0.10cm of s1] {4 evolvable harness slots};

\node[gepa, right=1.35cm of s1] (g1) {GEPA};
\node[gepa, right=1.35cm of s2] (g2) {GEPA};
\node[gepa, right=1.35cm of s3] (g3) {GEPA};
\node[gepa, right=1.35cm of s4] (g4) {GEPA};
\node[lbl, align=center] at ($(g1)+(0,0.66)$) {per-slot\\reflective evolution};

\draw[arr] (s1) -- (g1);
\draw[arr] (s2) -- (g2);
\draw[arr] (s3) -- (g3);
\draw[arr] (s4) -- (g4);
\draw[->, thick, blue!45!black, dashed] (g1.east) to[out=0,in=90] (g2.north);
\draw[->, thick, blue!45!black, dashed] (g2.east) to[out=0,in=90] (g3.north);
\draw[->, thick, blue!45!black, dashed] (g3.east) to[out=0,in=90] (g4.north);

\node[inj, right=1.5cm of g1, yshift=-0.32cm] (frz) {freeze \& render\\single string $H$};
\node[inj, below=0.5cm of frz] (inj) {\textbf{single fixed}\\\textbf{injection point}};
\draw[arr] (g1.east) -- ++(0.35,0) |- (frz.west);
\draw[arr] (g2.east) -- ++(0.20,0) |- (frz.west);
\draw[arr] (g3.east) -- ++(0.20,0) |- (frz.west);
\draw[arr] (g4.east) -- ++(0.35,0) |- (frz.west);
\draw[arr] (frz) -- (inj);

\node[inj, right=1.3cm of frz, yshift=-0.7cm, fill=gray!18] (agent) {frozen $M$\\ReAct rollout\\(held-out test)};
\draw[arr] (frz.east) to[out=0,in=130] (agent.north west);
\draw[arr] (inj.east) to[out=0,in=210] (agent.south west);

\node[draw=red!50!black, rounded corners, fill=red!4, font=\scriptsize, align=center]
  (iso) at ($(g4.south)+(0,-0.85)$) {iso-budget $B$ split across slots:\\$B/4$ per slot $<$ accept-floor $\Rightarrow$ all freeze empty};
\draw[->, red!55!black, dashed] (iso.north) -- ($(g4.south)+(0,-0.02)$);

\node[cred] (loo) at ($(s4.south)+(0.9,-2.25)$) {\textbf{\LOO}: evolve all-but-one slot\\$\Rightarrow$ removal cost (\emph{necessity})};
\node[cred, right=0.7cm of loo] (loi) {\textbf{\LOI}: evolve one slot alone\\$\Rightarrow$ marginal value (\emph{sufficiency})};
\draw[arr] (agent.south) to[out=270,in=0] (loi.east);
\draw[arr, dashed, gray!55!black] (loi.west) -- (loo.east);
\node[lbl, align=center, below=0.08cm of loo.south east, xshift=1.0cm]
  {credit-assignment map: held-out success, paired McNemar};

\end{tikzpicture}
\caption{\textbf{\method architecture.} The harness is four named slots
($c_1$ role, $c_2$ strategy, $c_3$ format, $c_4$ control). Coordinate ascent
evolves each slot in turn with the \emph{same} reflective optimizer (GEPA),
freezing earlier slots and seeding later ones (dashed arrows); the per-slot
budgets sum to the flat method's budget $B$ (iso-budget). The selected slots are
frozen, rendered into a single string, and injected at the \emph{single} fixed
injection point of the frozen-model ReAct rollout. The leave-one-in (\LOI) and
leave-one-out (\LOO) probes evolve one slot alone or all-but-one to produce a
held-out credit-assignment map. The red annotation marks the failure analyzed in
Section~\ref{sec:analysis}: a four-way split puts $B/4$ below the optimizer's
accept-and-rescore floor, so every slot freezes empty.}
\label{fig:arch}
\end{figure}

\method has three parts: (i) the four-slot harness of Section~\ref{sec:problem},
(ii) a coordinate-ascent procedure that evolves the slots one at a time with an
existing reflective optimizer under an iso-budget, and (iii) the \LOI/\LOO\
credit-assignment protocol. The architecture is shown in
Figure~\ref{fig:arch}.

\subsection{Architecture and the single injection point}

The defining design choice is that the harness is the \emph{only} thing that
varies. The four slots are rendered, per Eq.~\eqref{eq:render}, into a single
string that is injected at the exact point every comparison method uses---the
strategy preamble of the ReAct loop. The backbone $M$, the decoding temperature
($0$), the environment, the action parser, and the success verifier are identical
across stock, flat-string evolution, and \method. Consequently the stock harness
(all slots empty) reproduces the stock ReAct preamble byte-for-byte, and any
measured difference is attributable solely to injected text content and structure,
never to a changed rollout path. This is what licenses the credit-assignment
interpretation: a slot's \LOI\ effect is a clean text intervention.

\subsection{Coordinate-ascent evolution under an iso-budget}

We evolve the four slots by coordinate ascent (Algorithm~\ref{alg:harnessevo}).
The optimizer is the reflective prompt-evolution loop of GEPA
\citep{agrawal2025gepa}, reused unchanged: it maintains a pool of candidate
strings, selects a parent from the instance Pareto frontier, mutates it by
reflecting on recent rollout traces, accepts the child if it does not lose on a
training minibatch, and re-scores accepted children on the full validation set,
returning the best-validation candidate. We run this optimizer once per slot. When
evolving slot $c_k$, all earlier slots are \emph{frozen} at their already-evolved
text and all later slots are held at their empty seed; the optimizer's internal
empty seed candidate therefore corresponds exactly to ``the current harness with
slot $c_k$ neutral,'' the correct baseline for that coordinate step. A slot-scoped
reflection meta-prompt instructs the reflection model to improve \emph{only} slot
$c_k$ and shows the other slots as fixed context, so the optimizer does not
duplicate or fight content that lives in another slot.

The total rollout budget is capped to exactly equal the budget given to flat
single-string evolution (iso-budget): we split $B$ across the evolved slots, and
because the optimizer counts rollouts internally and stops at its per-slot cap, the
sum of per-slot budgets equals $B$ by construction. The rollout path is, again,
byte-identical to flat evolution; only the content and structure of the injected
text differ.

\begin{algorithm}[t]
\caption{\method coordinate-ascent harness evolution (iso-budget)}
\label{alg:harnessevo}
\DontPrintSemicolon
\KwIn{train set $T$, validation set $V$, frozen model $M$, slot order
$(c_1,\dots,c_m)$, total budget $B$, reflective optimizer \textsc{Gepa}}
\KwOut{evolved harness $H_{\mathrm{full}}$ and a per-slot history}
$H \leftarrow H_0$ \tcp*{stock: all slots empty}
$(b_1,\dots,b_m) \leftarrow \textsc{SplitEvenly}(B, m)$ \tcp*{per-slot budgets, $\sum_k b_k = B$}
\For{$k \leftarrow 1$ \KwTo $m$}{
  $\textit{ctx} \leftarrow$ render of all slots $\neq c_k$ as fixed context\;
  $\textit{tmpl}_k \leftarrow$ slot-scoped reflection meta-prompt for $c_k$ with $\textit{ctx}$ baked in\;
  define $\mathrm{roll}_k(\textit{text}, \tau) := \mathrm{rollout}\big(M, \mathrm{render}(H[c_k{\leftarrow}\textit{text}]), \tau\big)$\;
  $(\textit{best}_k,\, \textit{val}_k,\, \textit{info}_k) \leftarrow
     \textsc{Gepa}\big(T, V, M;\ \text{budget } b_k,\ \mathrm{roll}_k,\ \textit{tmpl}_k\big)$\;
  $H \leftarrow H[c_k \leftarrow \textit{best}_k]$ \tcp*{freeze val-selected best into slot $c_k$}
}
\Return $H,\ \text{history}$\;
\end{algorithm}

\subsection{The leave-one-in / leave-one-out protocol}

To produce the credit-assignment map, we instantiate the same evolution machinery
in two restricted modes. A \emph{leave-one-in} run for slot $c_k$ evolves only that
slot (all others fixed at the empty seed) and measures $\Delta^{\mathrm{LOI}}_k$,
the marginal value of $c_k$ alone over stock. A \emph{leave-one-out} run for slot
$c_k$ evolves all slots except $c_k$, at the same total budget $B$ as the full
method, and measures $\Delta^{\mathrm{LOO}}_k$, how much the full harness loses
without $c_k$. The two probes are complementary: \LOI\ asks ``is this slot
sufficient?'' and \LOO\ asks ``is this slot necessary?''. Each evolved harness is
frozen and evaluated once on the held-out test set under the shared rollout; we
report bootstrap confidence intervals on the test success rate and paired McNemar
tests against the relevant reference (stock for \LOI, full for \LOO). Single-slot
\LOI\ runs use budget $B/2$ by default---half the four-way split's total---which,
as Section~\ref{sec:analysis} shows, is already above the optimizer's accept-floor
and is part of the reason the targeted runs succeed where the split fails.

\section{Experimental Setup}
\label{sec:setup}

\paragraph{Backbone.} All experiments use a single \emph{frozen} Qwen2.5-7B model
\citep{yang2025qwen3} as both the task policy and the reflection model, with no
weight updates anywhere (the fairness iron law: the same model that acts also
reflects). Decoding is deterministic ($\text{temperature}{=}0$) for rollouts; the
reflection step samples mutations at higher temperature, as in
GEPA~\citep{agrawal2025gepa}.

\paragraph{Benchmarks and splits.} On \textbf{ALFWorld}
\citep{shridhar2021alfworld} we draw the train and validation tasks from a balanced
30-task development manifest and evaluate on a \emph{disjoint} held-out test
manifest of $n=134$ tasks (zero leakage by construction). The optimizer sees
$|T|=30$ training and a small held-out validation set for candidate selection;
episodes run up to $40$ steps with environment-verified binary success. On
\textbf{WebShop} \citep{yao2022webshop} we use a held-out test set of $n=80$
sessions with the benchmark's dense $[0,1]$ attribute-match score; train and
validation sessions are disjoint. Test sets are fixed across all methods.

\paragraph{Budgets.} The primary arm fixes an iso-budget of $B=64$ rollouts
(\emph{B64}): stock spends $\approx 0$, flat-string evolution spends the budget
on one string, and \method splits it across four slots ($16$ per slot). A
corroboration arm uses a higher budget, \emph{B120} ($B=120$, with a smaller
validation set and minibatch), to check robustness of the direction. For the
budget-allocation study we additionally run single-slot \LOI\ at $B/2=32$ rollouts
and an all-to-control run at the full $64$ rollouts.

\paragraph{Methods.} We compare: \textbf{Stock} (empty harness $H_0$);
\textbf{GEPA-flat} (flat single-string reflective evolution at budget $B$, the
principal foil); \textbf{\method-full} (coordinate ascent over all four slots at
budget $B$); the four \textbf{\LOI} runs (one per slot); the four \textbf{\LOO}
runs (all-but-one per slot); and the budget-allocation arms (\textbf{all-to-c4} at
$32$ and $64$). All methods share the identical rollout path.

\paragraph{Metrics and statistics.} The primary metric is held-out success rate
(ALFWorld, binary, single-greedy) or dense score (WebShop). We report
$10{,}000$-resample bootstrap $95\%$ confidence intervals on each rate, and
\emph{paired} significance tests on the per-task held-out outcomes: McNemar's test
for the binary ALFWorld metric (reporting discordant counts and $p$) and a paired
bootstrap on the WebShop dense score. Following the journal-style presentation
conventions we adopt, we report every number with an interval and a paired test,
label non-significant results ``ns,'' and \textbf{bold the true best in each
column even when it is a baseline}. We use $*,**,***$ for $p<0.05,0.01,0.001$.

\section{Results}
\label{sec:results}

\subsection{Main comparison: the full method ties}

Table~\ref{tab:main} reports the primary iso-budget ($B64$) comparison on ALFWorld
and the cross-benchmark WebShop arm. On the binary success metric, \method-full
reaches $0.657$, against $0.642$ for both the stock harness and flat-string
evolution. Neither difference is significant: McNemar gives $\Delta=+0.015$,
$p=0.617$ versus stock ($3$ vs $1$ discordant) and $\Delta=+0.015$, $p=0.480$
versus GEPA-flat ($2$ vs $0$ discordant); flat-string evolution itself ties stock
exactly ($\Delta=0.000$, $p=0.617$). \emph{All three methods tie.} We bold the
true best per column; because the three are statistically indistinguishable, the
honest reading is a three-way tie, and we mark the baselines as co-best rather than
manufacture a win for the structured method. On WebShop the same pattern holds at
the level of dense score: $0.518$ (stock) / $0.539$ (GEPA) / $0.545$ (\method),
with \method versus stock $\Delta=+0.027$, paired-bootstrap $p=0.054$ (ns), and
every slot frozen empty (Section~\ref{sec:analysis}). A reader stopping here would
conclude that structured harness evolution does nothing; the per-component map
shows why that conclusion is incomplete.

\begin{table}[t]
\centering
\small
\caption{\textbf{Main iso-budget comparison (the full method ties).} ALFWorld
held-out success ($n{=}134$, single-greedy, $10$k-bootstrap $95\%$ CI) at budget
$B{=}64$, and the cross-benchmark WebShop dense score ($n{=}80$). Paired McNemar
(ALFWorld) / paired bootstrap (WebShop) against \emph{Stock}; ns $=$ not
significant. \textbf{Bold} marks the true best per column; the three methods are
statistically indistinguishable on each benchmark, so the baselines are co-best
and \method does \emph{not} win.}
\label{tab:main}
\begin{tabularx}{\linewidth}{@{}l c c c X@{}}
\toprule
& \multicolumn{3}{c}{\textbf{ALFWorld} ($n{=}134$, binary success)} & \textbf{WebShop} \\
\cmidrule(lr){2-4}
Method & Success [95\% CI] & $\Delta$ vs Stock & McNemar $p$ & score [95\% CI] \\
\midrule
Stock        & \textbf{0.642} [0.560, 0.724] & ---     & ---          & \textbf{0.518} [0.447, 0.590] \\
GEPA-flat    & \textbf{0.642} [0.560, 0.724] & $+0.000$ & $0.617$ ns   & 0.539 [0.468, 0.609] \\
\method-full & 0.657 [0.575, 0.739]          & $+0.015$ & $0.617$ ns   & \textbf{0.545} [0.476, 0.613] \\
\bottomrule
\end{tabularx}

\vspace{0.4em}
{\footnotesize \emph{Reading.} On ALFWorld the three are a tie (also
\method vs GEPA-flat: $\Delta{=}+0.015$, $p{=}0.480$ ns). On WebShop \method vs
Stock is $\Delta{=}+0.027$, paired-bootstrap $p{=}0.054$ (ns). Bolding follows the
honesty rule: the best is a tie, and on ALFWorld a baseline is (co-)best.}
\end{table}

\subsection{Per-component credit-assignment map}

\begin{figure}[t]
  \centering
  \IfFileExists{figures/fig_components.pdf}{%
    \includegraphics[width=0.96\linewidth]{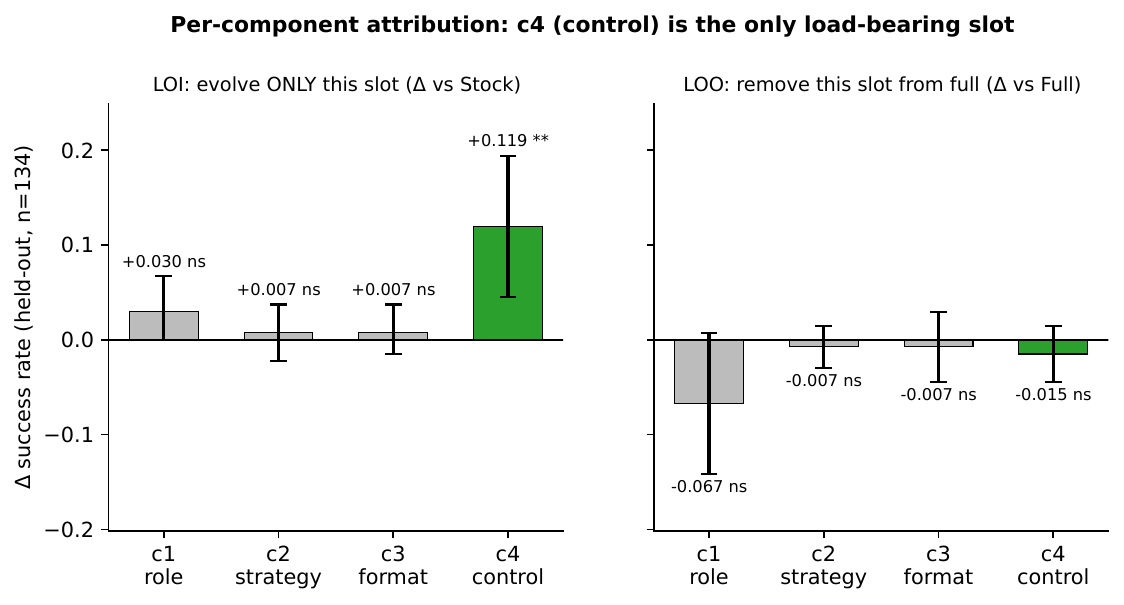}%
  }{\fbox{\parbox[c][3cm][c]{0.9\linewidth}{\centering\itshape figures/fig\_components.pdf not found.}}}
  \caption{\textbf{Per-component attribution.} Left: leave-one-in held-out gain
    over stock per slot---only \cctrl\ (control) is load-bearing ($+0.119$, $**$).
    Right: leave-one-out change relative to the full harness---no single removal is
    significant, because the full harness froze nearly empty. Bars are
    $95\%$ bootstrap intervals on the paired held-out difference ($n{=}134$).}
  \label{fig:components}
\end{figure}

Table~\ref{tab:loi} reports the \LOI\ and \LOO\ map on ALFWorld at $B64$; it is the
analytical heart of the paper and is visualized in
Figures~\ref{fig:teaser}~and~\ref{fig:components}. The marginal value is
overwhelmingly concentrated in the reflection/control slot \cctrl: evolving it
alone lifts held-out success by $\Delta^{\mathrm{LOI}}_{\cctrl}=+0.119$
($p=0.0046$, $**$; $22$ of the $28$ discordant tasks flip toward the evolved
control), reaching $0.761$ success. The other three slots are individually null:
$\Delta^{\mathrm{LOI}}_{\croleone}=+0.030$ (ns), $\Delta^{\mathrm{LOI}}_{\cstrat}=+0.007$
(ns), $\Delta^{\mathrm{LOI}}_{\cfmt}=+0.007$ (ns). The leave-one-out side is
consistent with a single load-bearing slot: removing any one slot from the
(near-empty) full harness changes held-out success by at most $0.067$ and never
significantly ($\Delta^{\mathrm{LOO}}_{\croleone}=-0.067$,
$\Delta^{\mathrm{LOO}}_{\cstrat}=-0.007$, $\Delta^{\mathrm{LOO}}_{\cfmt}=-0.007$,
$\Delta^{\mathrm{LOO}}_{\cctrl}=-0.015$, all ns)---unsurprising, since the full
harness froze almost entirely empty and there is little to remove. The decisive
quantity is the comparison of the full gain to the sum of single-slot gains: the
full harness improves over stock by only $+0.015$, while the \LOI\ gains sum to
$+0.164$. The system is strongly \emph{sub-additive}; the structured method
captures roughly one-tenth of what its components individually offer.

\begin{table}[t]
\centering
\small
\caption{\textbf{Per-component credit-assignment map on ALFWorld} ($B64$,
$n{=}134$, $10$k-bootstrap $95\%$ CI). \LOI: evolve only this slot, $\Delta$ vs
\emph{Stock}. \LOO: evolve all-but-this-slot, $\Delta$ vs \emph{\method-full}.
McNemar $p$; $**\,p{<}0.01$. The value is localized to \cctrl\ (control); the
others are individually null. Full gain ($+0.015$) $\ll \sum\!\LOI$ ($+0.164$):
strongly \emph{sub-additive}.}
\label{tab:loi}
\begin{tabularx}{\linewidth}{@{}llcccc@{}}
\toprule
Slot & Component & $\Delta^{\mathrm{LOI}}$ [95\% CI] & $p$ & $\Delta^{\mathrm{LOO}}$ [95\% CI] & $p$ \\
\midrule
$c_1$ & role/persona            & $+0.030$ [$+0.00,+0.07$] & 0.221 ns & $-0.067$ [$-0.14,+0.01$] & 0.124 ns \\
$c_2$ & task-strategy           & $+0.007$ [$-0.02,+0.04$] & 1.000 ns & $-0.007$ [$-0.03,+0.01$] & 1.000 ns \\
$c_3$ & tool/format-rules       & $+0.007$ [$-0.01,+0.04$] & 1.000 ns & $-0.007$ [$-0.04,+0.03$] & 1.000 ns \\
$c_4$ & reflection/control      & $\mathbf{+0.119}$ [$+0.04,+0.19$] & $\mathbf{0.0046}$\,** & $-0.015$ [$-0.04,+0.01$] & 0.617 ns \\
\midrule
\multicolumn{2}{@{}l}{\emph{full} ($H_{\mathrm{full}}$ vs Stock)} & \multicolumn{2}{l}{$+0.015$ (ns)} & \multicolumn{2}{r}{$\textstyle\sum\!\LOI = +0.164$} \\
\bottomrule
\end{tabularx}
\end{table}

\subsection{Budget corroboration (B120) and budget allocation}

The higher-budget arm (\emph{B120}) corroborates the direction while underscoring
that magnitudes are noisy at this sample size. There, stock measures higher
($0.672$), flat evolution gives $0.642$, and \method-full gives $0.627$; all
pairwise differences are ns. The control slot remains the most positive single-slot
direction ($\Delta^{\mathrm{LOI}}_{\cctrl}=+0.052$, ns here---non-significant
because B120 stock is higher and the evolved control validated weaker), and the
system is sub-additive again (full $-0.045$ vs $\sum\!\LOI=+0.052$). We treat B120
as a robustness check on the \emph{ordering} of slots, not as an independent
significant gain.

\begin{figure}[t]
  \centering
  \IfFileExists{figures/fig_budget.pdf}{%
    \includegraphics[width=0.98\linewidth]{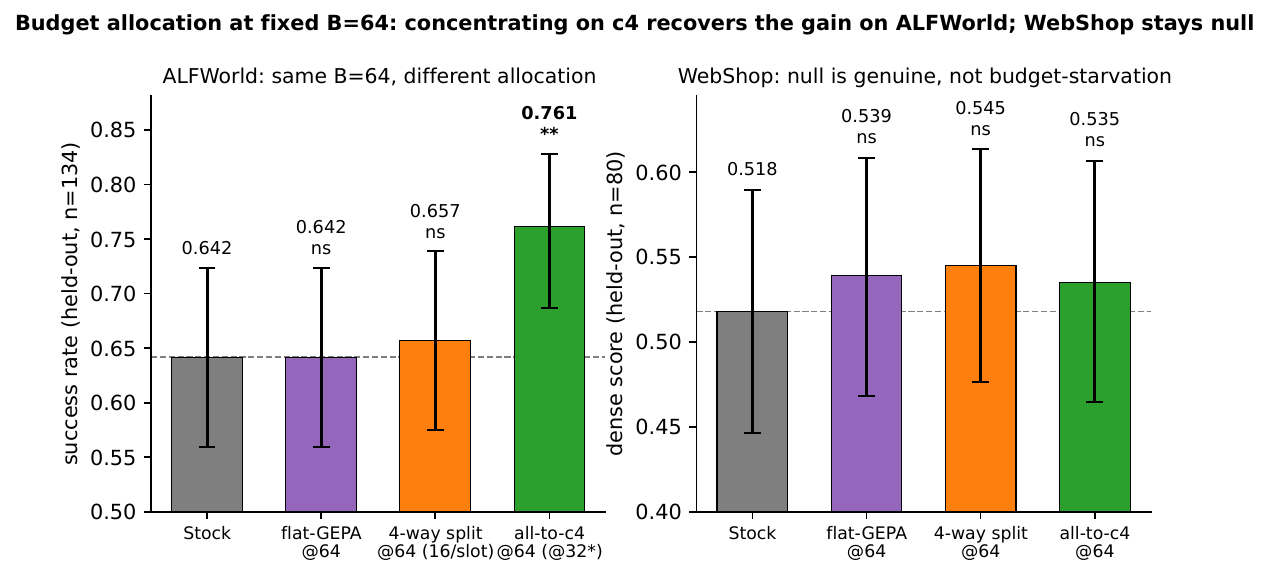}%
  }{\fbox{\parbox[c][3cm][c]{0.9\linewidth}{\centering\itshape figures/fig\_budget.pdf not found.}}}
  \caption{\textbf{Budget allocation.} Left (ALFWorld, $n{=}134$): at the same
    budget $B{=}64$, concentrating on the control slot (all-to-$c_4$, point shown at
    the significant @32 replication, $0.761$, $**$) far exceeds the four-way split
    ($0.657$, ns) and flat evolution ($0.642$, ns). Right (WebShop, $n{=}80$): the
    null is \emph{genuine}, not budget starvation---even all-to-$c_4$@64 stays at
    $0.535$ (ns), because the control slot froze empty for lack of a recurrent,
    verbalizable control failure to fix.}
  \label{fig:budget}
\end{figure}

The budget-allocation result is the practical payoff and we report it with full
honesty. Table~\ref{tab:budget} and Figure~\ref{fig:budget} compare, at fixed
total budget, the uniform four-way split against concentrating the budget on the
high-credit control slot. The four-way split ($16$/slot) reaches $0.657$ (ns) and
flat evolution reaches $0.642$ (ns). Concentrating on \cctrl\ recovers a large
gain: the control-only \LOI\ at \emph{half} the split's budget ($32$ rollouts)
reaches $0.761$, a $+0.119$ gain over stock that is significant ($p=0.0046$, $22$
vs $6$ discordant); the all-to-control run at the \emph{full} budget ($64$
rollouts) reaches $0.724$, a $+0.082$ gain that is positive and clearly above the
split's $0.657$ but borderline by McNemar ($21$ vs $10$ discordant, $p=0.0725$,
ns). We do not overclaim significance at $64$: the direction is robust (both
targeted runs beat the split, both exceed it by a wide margin), but the exact
magnitude fluctuates with evolution and decoding stochasticity, so the
defensible claim is that targeted single-slot evolution recovers a $+0.08$ to
$+0.12$ held-out gain that the uniform split discards.

\begin{table}[t]
\centering
\small
\caption{\textbf{Budget allocation at fixed budget on ALFWorld} ($n{=}134$,
$10$k-bootstrap $95\%$ CI, McNemar vs \emph{Stock}). Concentrating the budget on
the high-credit control slot recovers the gain the uniform four-way split
discards. We report \emph{both} the significant half-budget replication (@32) and
the borderline full-budget replication (@64); \textbf{bold} marks the true best.}
\label{tab:budget}
\begin{tabularx}{\linewidth}{@{}lccclX@{}}
\toprule
Allocation & Budget & Success [95\% CI] & $\Delta$ vs Stock & McNemar ($p$; wins) \\
\midrule
Stock (no search)        & $0$           & 0.642 [0.560, 0.724] & ---     & --- \\
GEPA-flat (one string)   & $64$          & 0.642 [0.560, 0.724] & $+0.000$ & $0.617$ ns ($2{:}2$) \\
4-way split (16/slot)    & $64$          & 0.657 [0.575, 0.739] & $+0.015$ & $0.617$ ns ($3{:}1$) \\
all-to-$c_4$ (full)      & $64$          & 0.724 [0.687, 0.828]\textsuperscript{$\dagger$} & $+0.082$ & $0.0725$ ns ($21{:}10$) \\
all-to-$c_4$ (half)      & $32$          & \textbf{0.761} [0.687, 0.828] & $\mathbf{+0.119}$ & $\mathbf{0.0046}$\,** ($22{:}6$) \\
\bottomrule
\end{tabularx}

\vspace{0.3em}
{\footnotesize \textsuperscript{$\dagger$}CI for all-to-$c_4$@64 ($0.724$) overlaps
that of @32; the point estimate is lower and only borderline-significant. We
deliberately do not claim significance at the full budget.}
\end{table}

\section{Analysis: Credit Assignment and the Budget-Splitting Trap}
\label{sec:analysis}

\subsection{The value is localized to control}

The \LOI\ map (Table~\ref{tab:loi}, Figure~\ref{fig:components}) makes a clean
statement: of the four harness components, only reflection/control carries
optimization value on ALFWorld, and it carries essentially all of it
($\Delta^{\mathrm{LOI}}_{\cctrl}=+0.119$, the sum over the other three is $+0.044$
and individually each is ns). This is not an artifact of the optimizer failing on
the other slots---the role and strategy slots \emph{were} explored and accepted
candidates in the targeted setting (their pools grew), the candidates simply did
not transfer to held-out success. The asymmetry is semantic: the frozen 7B agent
already knows how to phrase actions (so the format slot \cfmt\ adds nothing) and
already adopts a reasonable persona and plan (so \croleone\ and \cstrat\ add
nothing), but it lacks \emph{loop discipline}---it repeats dead-end actions, fails
to open closed receptacles before taking, and stops prematurely---and the control
slot supplies exactly that. The audited control text that produced the gain is:
``\emph{skip a repeated action that returned `Nothing happens'; open closed
receptacles before taking; confirm the object is held before placing; stop once
the goal is satisfied.}'' It is a compact, general, environment-grounded
correction, not a task-specific trick.

\subsection{Sub-additivity and the accept-and-rescore floor}

Why does the full method capture only $+0.015$ when its components offer $+0.164$?
The mechanism is a structural property of the reflective optimizer interacting with
budget-splitting. Each optimizer iteration spends $2\times\text{minibatch}$
rollouts scoring a parent and a candidate child; \emph{only on acceptance} does it
spend an additional block of rollouts re-scoring the accepted child on the full
validation set before admitting it to the pool. There is therefore a minimum
budget---an \emph{accept-and-rescore floor}---below which the optimizer cannot
accept and freeze even a single mutation: it must afford at least one
parent/child minibatch comparison \emph{and} the subsequent full-validation
rescore. With the B64 configuration (minibatch $3$, validation $\approx 10$), an
accept-bearing iteration costs about $2{\times}3 + 10 = 16$ rollouts---exactly the
$16$ rollouts a four-way split allots to each slot. The split therefore lands the
budget \emph{at or below} the floor for every slot, and the empirical consequence
is unambiguous: at $16$/slot, \emph{zero} of the four slots accepted any candidate
and all four froze at the empty seed (Table~\ref{tab:floor}). The structured
harness that emerged was byte-identical to stock, which is why \method-full ties
stock to within $+0.015$ of noise. By contrast, at $32$/slot three of four slots
accept a mutation but only one freezes non-empty text (the load-bearing control
slot), and at $60$/slot all four accept and freeze text. The floor is real, and
uniform splitting drives the per-slot budget straight through it.

\begin{table}[t]
\centering
\small
\caption{\textbf{The accept-and-rescore floor.} Number of slots (out of $4$) that
\emph{accepted} a mutation and that \emph{froze non-empty text}, by per-slot
budget. At $16$/slot (the B64 four-way split) every slot freezes empty; the floor
sits between $16$ and $32$ rollouts per slot.}
\label{tab:floor}
\begin{tabularx}{\linewidth}{@{}lXcc@{}}
\toprule
Per-slot budget & Arm & \#slots accepted & \#slots froze text \\
\midrule
$16$/slot & B64 four-way split (\method-full) & $0$ & $0$ \\
$32$/slot & B64 single-slot \LOI               & $3$ & $1$ \\
$30$/slot & B120 four-way split                & $3$ & $1$ \\
$60$/slot & B120 single-slot \LOI               & $4$ & $4$ \\
\bottomrule
\end{tabularx}
\end{table}

This yields the paper's sharpest practical statement: \emph{naive structured
evolution can be worse than a targeted single-slot run}, because decomposing the
harness divides the budget below the optimizer's working point. The remedy is not
more structure but better allocation.

\subsection{Budget concentration recovers the gain}

If the floor explains the failure, then concentrating the budget should fix it,
and it does (Table~\ref{tab:budget}, Figure~\ref{fig:budget}). Spending the entire
budget on the high-credit control slot keeps every rollout above the floor: the
control-only run accepts and freezes a substantive rule, and held-out success rises
to $0.761$ at half the split's budget ($+0.119$, $**$) and $0.724$ at the full
budget ($+0.082$, borderline). The half-budget run is, strikingly, the strongest
configuration we observe---more budget is not the issue; \emph{where} the budget
goes is. This is the prescription we draw: under a fixed budget, identify the
high-credit slot (via a cheap \LOI\ probe) and concentrate the budget there rather
than splitting it uniformly across named components. We report the @32 (significant)
and @64 (borderline) replications together to be candid that the magnitude is
noisy; the ordering---targeted $\gg$ split $\approx$ flat $\approx$ stock---is what
is robust.

\subsection{Task-type breakdown: where the control rule pays off}

\begin{figure}[t]
  \centering
  \IfFileExists{figures/fig_tasktype.pdf}{%
    \includegraphics[width=0.94\linewidth]{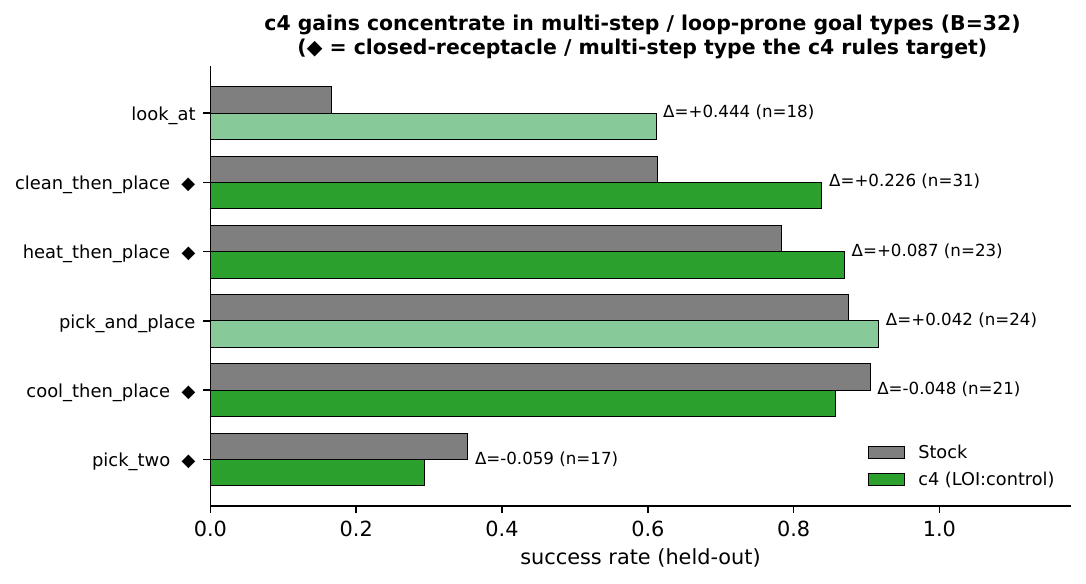}%
  }{\fbox{\parbox[c][3cm][c]{0.9\linewidth}{\centering\itshape figures/fig\_tasktype.pdf not found.}}}
  \caption{\textbf{Control-slot gain by ALFWorld task type} (control-only \LOI\ at
    $B{=}32$). Diamonds ($\blacklozenge$) mark closed-receptacle / multi-step types
    the control rules target. Gains concentrate in loop-prone, multi-step types
    (\texttt{look\_at}, \texttt{clean\_then\_place}); near-saturated or
    short-horizon types show little change. The effect tracks the rule's content.}
  \label{fig:tasktype}
\end{figure}

Figure~\ref{fig:tasktype} breaks the control-slot gain down by ALFWorld task type
(at the significant $B{=}32$ setting). The gains concentrate exactly where the
control rule's content predicts. The largest improvement is on
\texttt{look\_at\_obj\_in\_light} tasks ($\Delta=+0.444$, $n=18$): these require
the agent not to give up and to use the desk lamp while holding the object, a
classic premature-stopping / loop failure that the control rule directly
addresses. The next largest is on \texttt{clean\_then\_place} ($\Delta=+0.226$,
$n=31$), a multi-step, closed-receptacle type where ``open before taking'' and
``confirm holding before placing'' bite. Simple \texttt{pick\_and\_place}
($\Delta=+0.042$, $n=24$) and \texttt{heat\_then\_place} ($\Delta=+0.087$, $n=23$)
show small positive effects; the two types where the rule helps least or slightly
hurts---\texttt{cool\_then\_place} ($\Delta=-0.048$, $n=21$) and
\texttt{pick\_two\_obj\_and\_place} ($\Delta=-0.059$, $n=17$)---are where the
generic heuristic occasionally over-fires. Of the $22$ net task wins from the
control rule, half ($11$) fall in the closed-receptacle / multi-step types it was
written for. The gain is mechanistic, not diffuse.

\subsection{The WebShop null is genuine task-contingency, not starvation}

On WebShop every slot froze empty under \method-full, and the full method tied the
baselines (Table~\ref{tab:main}). A natural worry is that this null is just another
instance of the budget-splitting trap---perhaps WebShop too has a load-bearing
slot, starved by the four-way split. We rule this out directly. The all-to-control
run on WebShop, which hands the \emph{entire} budget to the control slot (well
above the accept floor), \emph{also} returns a null: dense score $0.535$ versus
$0.518$ for stock ($\Delta=+0.017$, paired-bootstrap $p=0.36$, ns), and the control
slot froze empty even with the concentrated budget. The optimizer had every
opportunity to accept a control rule and found none worth accepting. The WebShop
result is therefore a genuine property of the task, not an artifact of allocation:
WebShop's instruction-conditioned purchasing, with its dense attribute-match
reward, simply does not exhibit a recurrent, verbalizable control failure mode that
a general heuristic can repair. This is the boundary of the phenomenon---harness
optimization captures value precisely when the task has recurrent, verbalizable
control failures (dead-end actions, ordering constraints, premature stopping), and
is null otherwise.

\subsection{Grounded self-correction, not reward-hacking}

The value that survives is, on inspection, environment-grounded \emph{self-correction}
rather than reward-exploiting strategy. The control rule fixes real control errors
and generalizes across held-out task types (Figure~\ref{fig:tasktype}); it does not
encode a shortcut that games the binary success checker. Two observations reinforce
this. First, the gain transfers across task types it was not specifically tuned
on---the mark of a genuine correction rather than an over-fit. Second, on the
benchmark with the more gameable dense reward (WebShop), the optimizer declines to
fill any slot rather than fabricate a reward-chasing strategy. A harness optimizer
that captures grounded correction where it exists and abstains where it does not is
the desired behaviour in the presence of an imperfect environment reward; we
develop this as a constructive counterpoint to over-optimization concerns in
Section~\ref{sec:rel:goodhart}. Validation-to-test gaps were small or negative
throughout, indicating no over-fitting to the selection set.

\section{Discussion}
\label{sec:discussion}

\paragraph{Localize before you optimize.} The headline practical lesson is that
harness-optimization value is not diffuse across the scaffold; on our tasks it is
concentrated in a single component. This changes how a fixed optimization budget
should be spent. Rather than evolve the harness as one string (which cannot tell
you where the value is) or split a budget uniformly across named components (which,
as we show, can drive every component below the optimizer's accept-floor and
capture nothing), the budget should first be used cheaply to \emph{locate} the
high-credit component---a single-slot \LOI\ probe suffices---and then concentrated
there. Structured decomposition is valuable as a \emph{microscope} for this
localization, not as a drop-in method for a better agent.

\paragraph{The budget-splitting trap is a property of the optimizer, not the
harness.} The sub-additivity we observe is mechanistic: it follows from the
accept-and-rescore floor of the reflective optimizer. Any optimizer with a minimum
budget to accept and validate a candidate will exhibit some version of this trap
when its budget is divided across components below that minimum. The practical
implication generalizes beyond our specific optimizer: when distributing a fixed
search budget across parts of a system, account for the per-part minimum the search
procedure needs to make progress, or the division will silently waste the budget.

\paragraph{When does harness optimization help?} Our two benchmarks bracket the
answer. Harness optimization helped on ALFWorld and was null on WebShop, and the
audited slot text explains the difference: ALFWorld has recurrent, verbalizable
control failures that a compact general rule can fix, whereas WebShop does not.
This is a usable predictor. Before investing in harness optimization for a new
task, ask whether the frozen agent's failures are (i) recurrent across instances
and (ii) describable as a general control heuristic. If yes, a targeted control-slot
evolution is likely to pay off; if no, the effort will return a null, and---reassuringly---a
well-behaved optimizer will return that null rather than an over-fit.

\paragraph{The surviving value is the safe kind.} It is worth emphasizing that the
optimization value we localize is grounded self-correction, which is exactly the
kind of value one would want a black-box harness optimizer to capture: it fixes
real control errors, generalizes, and does not exploit the reward. On the benchmark
where the reward is most gameable, the method abstained. This is a small but
encouraging signal that careful, localized harness optimization need not amount to
reward-hacking.

\section{Conclusion}
\label{sec:conclusion}

We asked where, inside an LLM agent's harness, optimization value lives, and we
built a structured microscope to find out: \method, a four-slot decomposition
evolved by coordinate ascent under an iso-budget, paired with a leave-one-in /
leave-one-out credit-assignment protocol. The honest binary result is a tie: the
full structured method matches both the stock harness and flat single-string
evolution on held-out success ($0.657$ / $0.642$ / $0.642$, ns). The finding is
\emph{where} the value lives and \emph{how budget destroys it}. On ALFWorld the
value is localized almost entirely to the reflection/control slot (leave-one-in
$+0.119$, $p=0.0046$; role, strategy, format individually null). The structured
method ties not because the value is absent but because it is strongly
sub-additive: splitting an iso-budget of $64$ rollouts across four slots leaves
$16$ per slot, below the optimizer's accept-and-rescore floor, so every slot froze
empty and the load-bearing gain was never captured---the budget-splitting trap.
Concentrating the same budget on the high-credit slot recovers the discarded gain
(targeted control-only: $0.761$, $+0.119$, $p=0.0046$ at half the budget; $0.724$,
$+0.082$, borderline at the full budget---reported together, with no overclaim at
the larger budget). The phenomenon is task-contingent: on WebShop every slot froze
empty, a targeted control run stayed null, and the absence is a genuine property of
the task rather than starvation. The value that does survive is environment-grounded
self-correction, not reward-exploitation. We conclude that credit assignment and
budget concentration should precede any structured agent-evolution scheme:
\emph{localize first, then concentrate the budget on the one slot that matters.}

\paragraph{Limitations.} We name them plainly.
\emph{(1) Single frozen backbone.} All results use one frozen Qwen2.5-7B model; we
do not run a larger backbone. Whether the control-slot localization persists, shifts
to another slot, or vanishes at larger scale is the natural next study---we
explicitly flag backbone-scaling of the localization as future work and make
\emph{no} claim about it here.
\emph{(2) Two benchmarks, one of them null.} We study ALFWorld and WebShop; the
WebShop null is reported as a task-contingency boundary, not a weakness, but two
environments cannot map the full space of when harness optimization helps.
\emph{(3) One structured scheme.} We evolve slots by coordinate ascent with a
single reflective optimizer; other decompositions (different slot taxonomies, joint
rather than sequential evolution, a different base optimizer) might split value
differently, and our sub-additivity mechanism is tied to this optimizer's
accept-floor.
\emph{(4) Small evolve/validation splits and sample sizes.} The held-out sets are
modest ($n{=}134$ ALFWorld, $n{=}80$ WebShop) and the validation sets used for
candidate selection are small; this is why the budget-allocation magnitude is
significant at $@32$ but only borderline at $@64$, and why we make directional
rather than precise-magnitude claims there.

\paragraph{Future work.} Beyond backbone scaling, the most promising directions are
a cheap automatic \emph{slot-credit estimator} that allocates budget proportional
to estimated per-slot value (turning our manual prescription into a one-pass
allocator), and a taxonomy of tasks by their control-failure profile to predict, a
priori, whether targeted harness optimization will pay off.

\bibliographystyle{unsrtnat}
\bibliography{references,references_extra}

\appendix
\section{Frozen Slot Texts, Additional Arms, and Hyperparameters}
\label{sec:appendix}

\subsection{Audited frozen slot texts}
\label{app:frozen}

The localization claim rests on auditable text. We reproduce the control-slot
content that produced the significant ALFWorld gain (the $B{=}32$ control-only run,
$401$ characters, frozen after validation selection):

{\par\medskip\noindent\footnotesize\ttfamily\raggedright\sloppy
think: Before performing any action, ensure the previous step succeeded.\\
if action returned ``Nothing happens'' and is identical to the last action, skip
it and try another location.\\
open any closed receptacle before taking from it.\\
confirm the object is held before attempting to place it.\\
stop the sequence once the task is completed successfully or all relevant
locations are checked without success.\par\medskip}

The full-budget all-to-control run ($B{=}64$) froze a longer, pseudo-code variant of
the same rules ($1000$ characters; an \texttt{if}/\texttt{then} elaboration of
``skip repeated dead-end actions, open before taking, verify holding before
placing, stop when satisfied''). On \textbf{WebShop}, by contrast, the control slot
(and every other slot, under both the four-way split and the all-to-control run)
froze \emph{empty} ($0$ characters): the optimizer accepted no candidate. This
empty-versus-substantive contrast across benchmarks is the direct, auditable
signature of the task-contingency boundary in Section~\ref{sec:analysis}. The
role, strategy, and format slots on ALFWorld also froze empty in the iso-budget
full run ($0$ characters each), consistent with their null \LOI\ effects.

\subsection{B120 corroboration arm (full table)}
\label{app:b120}

Table~\ref{tab:b120} gives the higher-budget arm in full. Stock measures higher at
this configuration ($0.672$), flat evolution gives $0.642$, and \method-full gives
$0.627$; all pairwise McNemar tests are ns. The control slot remains the most
positive single-slot direction ($\Delta^{\mathrm{LOI}}_{\cctrl}=+0.052$, ns), and
the system is sub-additive (full $-0.045$ vs $\sum\!\LOI=+0.052$). We include this
arm as a robustness check on the \emph{ordering} of per-slot value, not as an
independent significant result.

\begin{table}[t]
\centering
\small
\caption{B120 corroboration arm on ALFWorld ($n{=}134$, $10$k-bootstrap CI).
Main methods (top) and per-component \LOI\ map (bottom). All differences ns;
the control direction remains the largest single-slot gain and the system is
sub-additive.}
\label{tab:b120}
\begin{tabularx}{\linewidth}{@{}lccX@{}}
\toprule
Method / Slot & Success / $\Delta^{\mathrm{LOI}}$ [95\% CI] & McNemar $p$ & Rollouts \\
\midrule
Stock        & 0.672 [0.590, 0.746] & ---        & $0$ \\
GEPA-flat    & 0.642 [0.560, 0.716] & $0.343$ ns & $120$ \\
\method-full & 0.627 [0.545, 0.709] & $0.377$ ns & $120$ \\
\midrule
$c_1$ role     & $+0.007$ [$-0.02,+0.04$] & 1.000 ns & $60$ \\
$c_2$ strategy & $+0.015$ [$-0.07,+0.10$] & 0.860 ns & $60$ \\
$c_3$ format   & $-0.022$ [$-0.07,+0.02$] & 0.546 ns & $60$ \\
$c_4$ control  & $+0.052$ [$-0.02,+0.13$] & 0.265 ns & $60$ \\
\bottomrule
\end{tabularx}
\end{table}

\subsection{Hyperparameters and protocol}
\label{app:hyper}

\begin{itemize}
\item \textbf{Backbone:} frozen Qwen2.5-7B, used as both task policy and reflection
model; rollouts deterministic ($\text{temperature}{=}0$); reflection mutations
sampled (temperature $0.9$), as in GEPA.
\item \textbf{ALFWorld:} train $|T|{=}30$ (balanced over task types from a 30-task
dev manifest), validation $\approx 10$ (B64) / $8$ (B120), test $n{=}134$ (disjoint
held-out manifest, zero leakage); max $40$ steps/episode; binary
environment-verified success.
\item \textbf{WebShop:} disjoint train/validation/test sessions, test $n{=}80$;
dense $[0,1]$ attribute-match score.
\item \textbf{Optimizer (GEPA):} instance-Pareto parent selection over a growing
candidate pool; reflective mutation on a train minibatch (size $3$ at B64, $2$ at
B120); improvement-or-equal accept gate for the binary-sparse reward; accepted
children re-scored on the full validation set; final $=$ best-validation candidate,
frozen. Iso-budget: per-slot budgets sum exactly to the flat method's budget.
\item \textbf{Budgets:} $B{=}64$ (primary), $B{=}120$ (corroboration); single-slot
\LOI\ at $B/2$; all-to-control at $32$ and $64$.
\item \textbf{Statistics:} $10{,}000$-resample bootstrap $95\%$ CIs; paired McNemar
(ALFWorld binary) / paired bootstrap (WebShop dense); ns labeled; $*/**/***$ for
$p<0.05/0.01/0.001$.
\item \textbf{Accept-floor diagnostic:} an accept-bearing GEPA iteration costs
$2\times\text{minibatch}+|V|$ rollouts; at B64 ($\text{minibatch}{=}3$, $|V|{\approx}10$)
this is $\approx16$, equal to the four-way split's $16$/slot, which is why the
split froze every slot empty (Table~\ref{tab:floor}).
\end{itemize}

\end{document}